%% file: paper.tex
\documentclass{article}

\usepackage{microtype}
\usepackage{graphicx}
\usepackage{subfigure}
\usepackage{booktabs} 
\usepackage{bera}
\usepackage{listings}
\usepackage{xcolor}
\usepackage{multicol}
\usepackage{filecontents}
\usepackage{lipsum}
\usepackage[hyphens]{url}
\usepackage[hidelinks]{hyperref}
\hypersetup{breaklinks=true}
\colorlet{punct}{red!60!black}
\definecolor{background}{HTML}{EEEEEE}
\definecolor{delim}{RGB}{20,105,176}
\colorlet{numb}{magenta!60!black}

\lstdefinelanguage{json}{
    basicstyle=\normalfont\ttfamily,
    showstringspaces=false,
    breaklines=true,
    frame=lines,
    backgroundcolor=\color{background},
    literate=
     *{0}{{{\color{numb}0}}}{1}
      {1}{{{\color{numb}1}}}{1}
      {2}{{{\color{numb}2}}}{1}
      {3}{{{\color{numb}3}}}{1}
      {4}{{{\color{numb}4}}}{1}
      {5}{{{\color{numb}5}}}{1}
      {6}{{{\color{numb}6}}}{1}
      {7}{{{\color{numb}7}}}{1}
      {8}{{{\color{numb}8}}}{1}
      {9}{{{\color{numb}9}}}{1}
      {:}{{{\color{punct}{:}}}}{1}
      {,}{{{\color{punct}{,}}}}{1}
      {\{}{{{\color{delim}{\{}}}}{1}
      {\}}{{{\color{delim}{\}}}}}{1}
      {[}{{{\color{delim}{[}}}}{1}
      {]}{{{\color{delim}{]}}}}{1},
}

\usepackage{hyperref}

\usepackage[accepted]{packages/sysml2019}

\sysmltitlerunning{Democratizing Production-Scale Distributed Deep Learning}

\begin{document}

\twocolumn[
\sysmltitle{Democratizing Production-Scale Distributed Deep Learning}




\begin{sysmlauthorlist}
\sysmlauthor{Minghuang Ma}{appl}
\sysmlauthor{Hadi Pouransari}{appl}
\sysmlauthor{Daniel Chao}{appl}
\sysmlauthor{Saurabh Adya}{appl}
\sysmlauthor{Santiago Akle Serrano}{appl}
\sysmlauthor{Yi Qin}{appl}
\sysmlauthor{Dan Gimnicher}{appl}
\sysmlauthor{Dominic Walsh}{appl}
\end{sysmlauthorlist}

\sysmlaffiliation{appl}{Apple Inc., Cupertino, California, USA}

\sysmlcorrespondingauthor{Minghuang Ma}{minghuang\_ma@apple.com}
\sysmlcorrespondingauthor{Hadi Pouransari}{mpouransari@apple.com}

\sysmlkeywords{Distributed Training, Kubernetes}

\vskip 0.3in

\input{text/abstract}
]



\printAffiliationsAndNotice{}  

\input{text/introduction}
\input{text/overview}
\input{text/implementation}
\input{text/distributed_training}
\input{text/case_studies}
\input{text/future_work}

\section*{Acknowledgements}
We would like to thank Anders Boesen Lindbo Larsen, Wenda Wang, Yin Zhou, Shreyas Saxena, Vinay Palakkode, Jort Gemmeke, Luciano Spinello for providing early feedback of the system. We would like to thank Michael Andrews, Jarrod Nettles for infrastructure support. We would like to thank Young Park, Niko Milonopoulos for project management. We would like to thank Mark Reid, Cheng Leong, Joelle Lam for many helpful discussions.



\nocite{langley00}

\Urlmuskip=0mu plus 1mu\relax
\bibliography{bib/sysml_paper}
\bibliographystyle{bib/sysml2019}





\end{document}

%% file: text/abstract.tex
\begin{abstract}\label{abstract}
  The interest and demand for training deep neural networks have been experiencing rapid growth, spanning a wide range of applications in both academia and industry. However, training them distributed and at scale remains difficult due to the complex ecosystem of tools and hardware involved. One consequence is that the responsibility of orchestrating these complex components is often left to one-off scripts and glue code customized for specific problems. To address these restrictions, we introduce \emph{Alchemist} - an internal service built at Apple from the ground up for \emph{easy}, \emph{fast}, and \emph{scalable} distributed training. We discuss its design, implementation, and examples of running different flavors of distributed training. We also present case studies of its internal adoption in the development of autonomous systems, where training times have been reduced by 10x to keep up with the ever-growing data collection.
\end{abstract}

%% file: text/introduction.tex
\section{Introduction}\label{sec:introduction}
In recent years, deep learning has become one of the most effective machine learning methodologies in many application domains including computer vision \cite{alexnet, vggnet, inceptionnet, resnet}, speech recognition \cite{deepspeech, deepspeech2, deepspeech3} and natural language processing \cite{rnnmt, googlenmt}. Despite the widespread adoption of deep neural networks (DNNs), training them quickly and at scale still poses significant \emph{computational}, \emph{algorithmic}, and \emph{engineering} challenges.

To address the computational challenge, the most direct way is to simply obtain better hardware. However, using a single state-of-the-art GPU to train the ResNet-50 model \cite{resnet} on the ImageNet dataset \cite{imagenet} can still take a few days. Despite the increase in computational power of GPUs and the development of powerful alternative hardware like TPUs \cite{tpu:paper}, the most common approach to reduce training time is to parallelize the computation across multiple GPUs hosted across multiple compute instances.

The algorithmic challenge, especially as distributed training is scaled out, lies in at least two aspects. Firstly, when large mini-batches are used, the model's ability to generalize tends to degrade \cite{largebatch:paper}. Secondly, as more compute nodes participate, the communication overhead grows. Luckily, these issues can all be mitigated. The first by carefully scheduling the changes in learning rate \cite{goyal2017accurate}, and the second, by adopting high performance computing (HPC\footnote{\url{http://research.baidu.com/bringing-hpc-techniques-deep-learning}}) techniques for efficient communication. Algorithmic advancements such as these have made distributed training tremendously successful, allowing researchers to train ImageNet in an hour \cite{goyal2017accurate} and even in just a few minutes \cite{akiba2017extremely,tencent4min}.

Last but not least, a real-world ML system requires significant infrastructure engineering in addition to ML code as mentioned in \cite{sculley2015hidden}. And the complexity of this infrastructure increases as it is leveraged by multiple teams of engineers, researchers, and data scientists. Each with different requirements and use cases. The engineering challenge can be further divided into the following:

\textbf{Interacting with compute resources.} We find two common patterns used in DNN training - interactive and batch. By interactive we mean the user logs into a remote shell to prototype and debug. By batch we mean the jobs are sent to a job queue where they are executed by the ML system. Interactive mode is usually preferred in the research and prototyping phase. On the other hand, batch mode is more suitable for large-scale parameter tuning and production pipelines. For both patterns, distributed training introduces additional overhead. Examples of challenges include management of IP addresses, starting and stopping processes, managing the synchronization of multiple processes possibly in multiple machines and integrating logs emanating from each process.

\textbf{Interacting with data resources.} Datasets tend to be shared by multiple training jobs, users, and teams. Given the write-once-read-many-times usage pattern, it is common to store and distribute the data using a distributed file system. The challenge becomes creating a performant and easy to access setup of this type. For example, Tensorflow provides an interface to implement custom plugins for different file systems, but we have found that a POSIX interface is often preferred by the users for its simplicity.

\textbf{Managing software and dependencies.} The software stack to train DNNs has become more complete than ever, ranging from low-level GPU libraries(CUDA, CUDNN) to high-level frameworks(Keras, Horovod \cite{horovod:paper}). Integrating these valuable software tools comes dependency management overhead. And in a remote and/or distributed environments, deploying the right software to the compute node is both complex and essential. Moreover, the software environment has to be the same on the cloud and the engineer's desk to simplify the development, debugging and code sharing.

\textbf{Availability of monitoring tools.} Monitoring the health and performance of a training job requires tools at the application level (e.g. Tensorboard, Tensorflow-timeline) as well as at the system level (e.g. to monitor CPU, memory, and GPU usage). Furthermore, distributed training requires us to also measure the communication between nodes in order to identify health issues and performance bottlenecks. Having tools for all of the above is essential for a dependable DNN development environment.

To address the above challenges, we discuss a system we built at Apple known as \emph{Alchemist}. Alchemist adopts a cloud-native architecture and is portable among private and public clouds. It supports multiple training frameworks like Tensorflow or PyTorch and multiple distributed training paradigms. The compute cluster is managed by, but not limited to, Kubernetes \footnote{\url{https://kubernetes.io}}. We chose a containerized workflow to ensure uniformity and repeatability of the software environment. In the following sections, we refer to engineers, researchers, and data scientists using Alchemist as \emph{users}.

\subsection{Related Work}
Multiple groups have developed systems to facilitate machine learning (ML) research and engineering workflows. Google's TensorFlow Extended (TFX) \cite{tfx:paper} is a ML platform that integrates all components of a ML pipeline and helps reduce time to production. Kubeflow \footnote{\url{https://www.kubeflow.org}} aims to provide a straightforward way to deploy open-source systems for ML over infrastructure managed by Kubernetes. Uber's Michelangelo \footnote{\url{http://eng.uber.com/michelangelo/}} is an internal ML-as-a-service platform designed to build and deploy ML related services. Twitter's DeepBird \footnote{\url{https://blog.twitter.com/engineering/en_us/topics/insights/2018/twittertensorflow.html}} is an end-to-end solution for training and serving deep learning models. Alchemist is similar in many aspects to these platforms, but with more focus on model training. Unlike Kubeflow's Kubernetes native approach, Alchemist is only using Kubernetes as a container orchestration platform.

There are many libraries and frameworks aimed at distributed training. Kubernetes's custom resource operators like \emph{tf-operator} and \emph{mpi-operator} have been integrated into Kubeflow. Horovod is a distributed training framework for Tensorflow, Keras, and PyTorch, which leverages the mature message passing interface (MPI) standard together with NCCL \footnote{\url{https://github.com/NVIDIA/nccl}} to provide efficient communication. BigDL \cite{wang2018bigdl} is a distributed deep learning library which allows users to write deep learning applications as standard Apache Spark programs. Ray \cite{moritz2017ray} provides a distributed framework with dynamic task scheduling to perform large-scale ML training jobs. Primitives for distributed training in Ray was introduced in \cite{bulatovflexible}. In addition, Tune \cite{liaw2018tune} is used to handle hyper-parameter tuning and model selection in conjunction with Ray. Alchemist is agnostic to the choice of distributed training framework and allows users to choose the tools most suitable for their needs.

%% file: text/overview.tex
\section{System Overview}\label{sec:overview}
\subsection{Parallelization Strategies}\label{sec:parallelization}
Since Alchemist's design is informed by various strategies in performing distributed stochastic gradient descent (SGD), we shall briefly review the possibilities below. Recall that each iteration of SGD consists of two phases: the forward and the backward passes. The forward pass uses the weights in the current iteration to compute the value of the activations and the loss function. These values are cached and used by the backward pass to compute gradients, which are subsequently applied to obtain the weights for the next iteration.

The strategies for distributing the SGD compute can be grouped into two main categories; those that are \emph{model} or \emph{data} parallel\footnote{See \cite{ben2018demystifying} for a more detailed discussion.}.

\textbf{Model parallel.}
This paradigm is suitable for very large models that cannot fit in the memory of a single compute node. In this approach, model weights are partitioned into groups, each of which are assigned to different compute nodes. It follows that completing a single SGD iteration requires intermediate values such as activations and gradients to be communicated across node boundaries, with the pattern of communication being dependent on the network topology and connectivity. Although model parallel strategies have conventionally been thought to exhibit limited scaling for convolutional neural networks(CNNs) that have spatially shared weights, works such as \cite{coates2013deep,ngiam2010tiled} have shown that they are promising.

\textbf{Data parallel.} In this paradigm, every compute node maintains a full copy of the model as well as its weights. Each node updates weights independently by performing the entirety of a single SGD iteration on a different subset of the data. Communication between nodes only happens when weight estimates are shared, which can be done either \emph{synchronously} or \emph{asynchronously}. In the synchronous case, gradients computed by each node is gathered and averaged before the weights are updated. In the asynchronous case, each node is allowed to update weights as soon as their gradients are ready without having to consult those computed by the others. The synchronous case typically exhibits better convergence while the asynchronous case typically exhibits lower communication costs.

Compared to the model parallel paradigm, the data parallel paradigm is both model agnostic and simpler to implement. This is not without its drawbacks; in particular, the amount of data processed for a single gradient estimate grows as the number of compute nodes participating grows. If unchecked this increase in the global batch-size can result in poor convergence \cite{goyal2017accurate,akiba2017extremely} as well as the degradation of the model's ability to generalize \cite{largebatch:paper}.

The patterns for SGD data communication can also be grouped into two main categories; those that are centralized and those that are not.

\textbf{Centralized communication.} In this pattern, one or a few compute nodes are designated as \emph{parameter servers} while the others are designated as \emph{workers}. Workers are responsible for computing gradients and publishing them to the parameter servers, while the parameter servers simply act as a hub where gradient estimates are gathered and broadcasted. An advantage of this pattern is that it is more resilient to the addition or removal of individual workers; this is especially helpful since machines do fail in practice.

\textbf{Decentralized communication.} Unlike the centralized pattern, there are no special roles and all nodes compute gradients and update weights. Communication occurs when gradient estimates are averaged, which typically proceeds through a ring all-reduce. Compared to the centralized pattern, this approach demonstrates better performance while being less fault-tolerant. Efficient implementation primitives have been developed over decades by the HPC community and are available in libraries like OpenMPI \footnote{\url{https://www.open-mpi.org}}, NCCL, and MLSL \footnote{\url{https://github.com/intel/MLSL}}. These primitives are then leveraged by higher level frameworks such as Horovod.

\subsection{Design Principles and Constraints}
Alchemist adheres to the following design principles with the objective of allowing users to easily leverage distributed optimization, for both research and production pipelines and do so efficiently.

\textbf{Easy distributed computing.} When writing distributed training code, users should not have to worry about the compute infrastructure, its orchestration, or its configuration. These complexities are the system's responsibility to abstract away. The users should be able to focus on their domain problem, provide code that defines the experiments, and be able to straightforwardly specify the runtime environment that they wish to work in.

\textbf{Portable architecture.} The system should be portable among private data centers and the public cloud. The system should be easily deployed into a different cloud in order to support legacy private compute clusters, isolate confidential product teams, isolate research and production environments, support multiple regions for fault tolerance, and to leverage additional GPU resources on public clouds.

\textbf{Full observability.} To enable a pleasant debugging experience and to facilitate the triaging of performance problems, the system should be fully observable. For example, it should be easy for users to leverage tools like Tensorboard to monitor the training progress. Furthermore, training task logs and compute metrics such as CPU, GPU, and memory usage should be easily accessible.

\textbf{Agnostic to training frameworks.} While there is a benefit in adopting a single training framework within a team or throughout an organization, users in practice choose different frameworks for different problems at different phases of a project. In response, the system should aim to enable and support different training frameworks.

\subsection{System Architecture}
\begin{figure}[h!]
\includegraphics[width=\columnwidth]{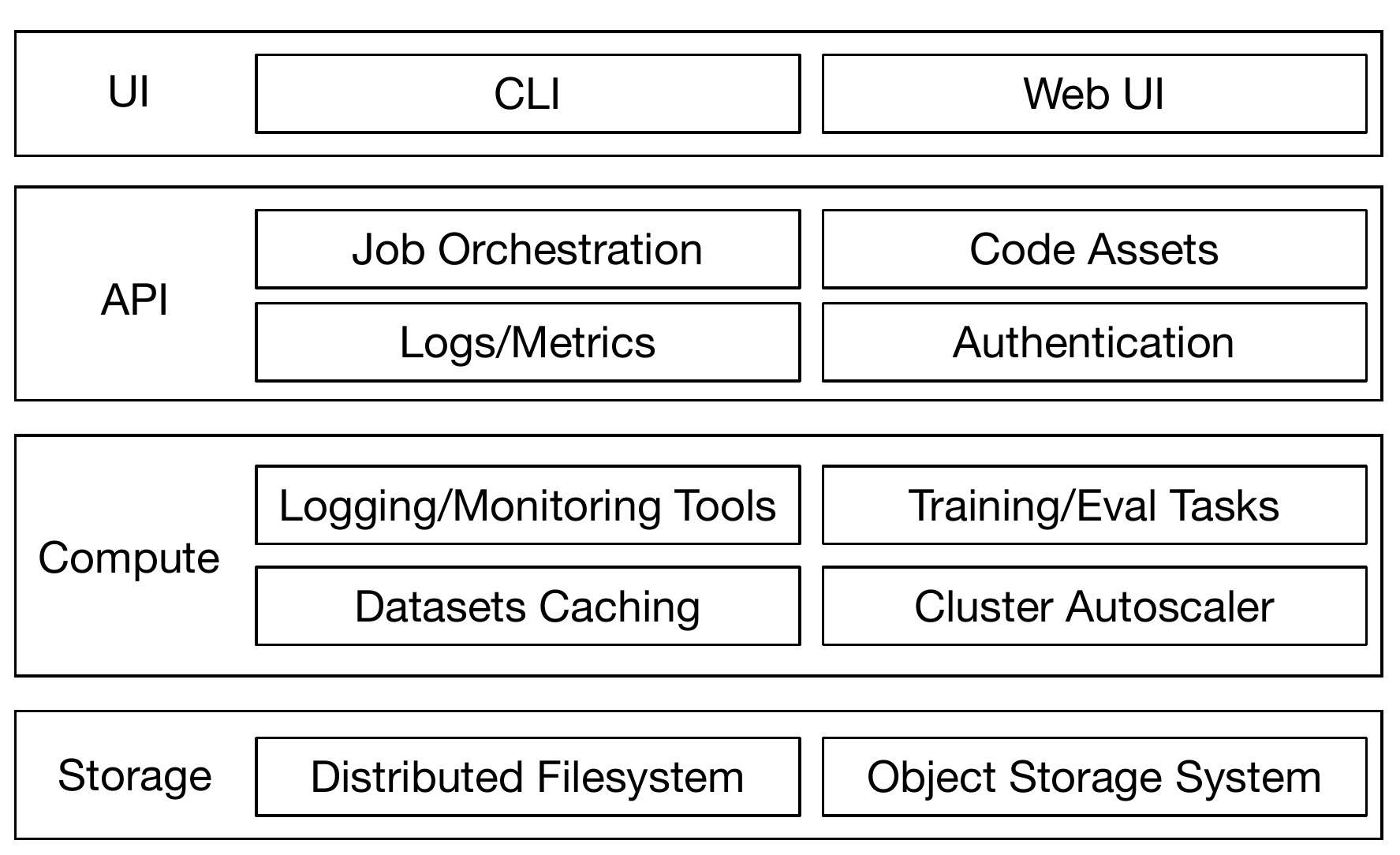}
\vspace*{-5mm}
\caption{System components fall into four layers. Layers are decoupled to simplify the support of new compute and storage platforms. For example, the compute layer components can be implemented using different container orchestration platforms.}
\vspace*{-2mm}
\label{fig:arch}
\end{figure}

Alchemist is designed with a layered micro-service architecture. This is illustrated in Figure \ref{fig:arch} and discussed below.

\textbf{Storage Layer.} This layer contains the distributed file systems and object storage systems that maintain the training and evaluation datasets, the trained model artifacts, and the code assets. Depending on the size and location of the datasets and annotations, there are many options ranging from using solely NFS, solely an object store, or a mixture of different storage systems. We discuss our choices of storage solutions in section \ref{sec:implementation}.

\textbf{Compute Layer.} This layer contains the compute-intensive workloads; in particular, the training and evaluation tasks orchestrated by the job scheduler. System monitoring and logging tools are preinstalled alongside to provide full observability. To improve compute resource utilization, a cluster auto-scaler watches the job queues and dynamically adjusts for the size of the cluster. Optionally, dedicated compute instances are deployed to provide dataset caching or to speed up data loading.

\textbf{API Layer.} This layer exposes services to users, which allows them to upload and browse the code assets, submit distributed jobs, and query the logs and metrics. Several API services are deployed behind a gateway that uses a central user authentication and authorization mechanism.

\begin{figure*}
\begin{center}
\includegraphics[width=0.9\textwidth]{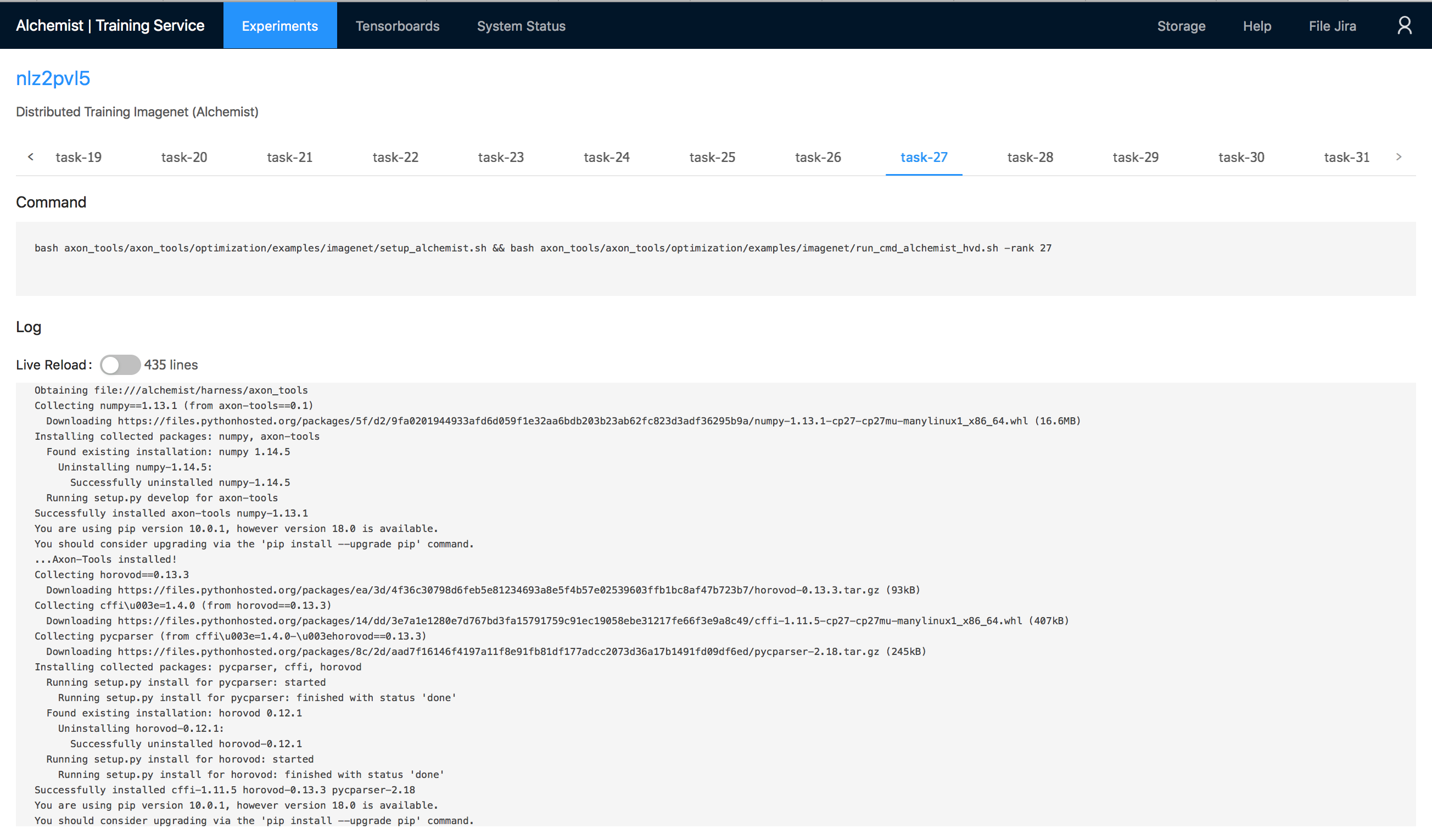}
\end{center}
\caption{Through the web UI, users can browse job histories, monitor job statuses, view their logs and resource utilization. The web UI also shows system health metrics, and provides a file browser to list datasets and model artifacts.}
\label{fig:ui}
\end{figure*}

\textbf{UI Layer.} This layer is composed of a command line interface (CLI) and a web UI. The CLI tool provides a practical way to submit and manage experiments. Most importantly, this enables the integration with CI/CD pipelines through scripting. The web UI allows the user to easily browse experiments, datasets, and artifacts. It also exposes the system's status and metrics to help with debugging and performance tuning. Figure \ref{fig:ui} shows an example of an 32-instance(256-GPU) distributed training experiment.

%% file: text/implementation.tex
\section{Implementation} \label{sec:implementation}
We describe the implementation of the key components of Alchemist below.

\subsection{Distributed job orchestration}
\begin{figure}[h!]
    \includegraphics[width=\columnwidth]{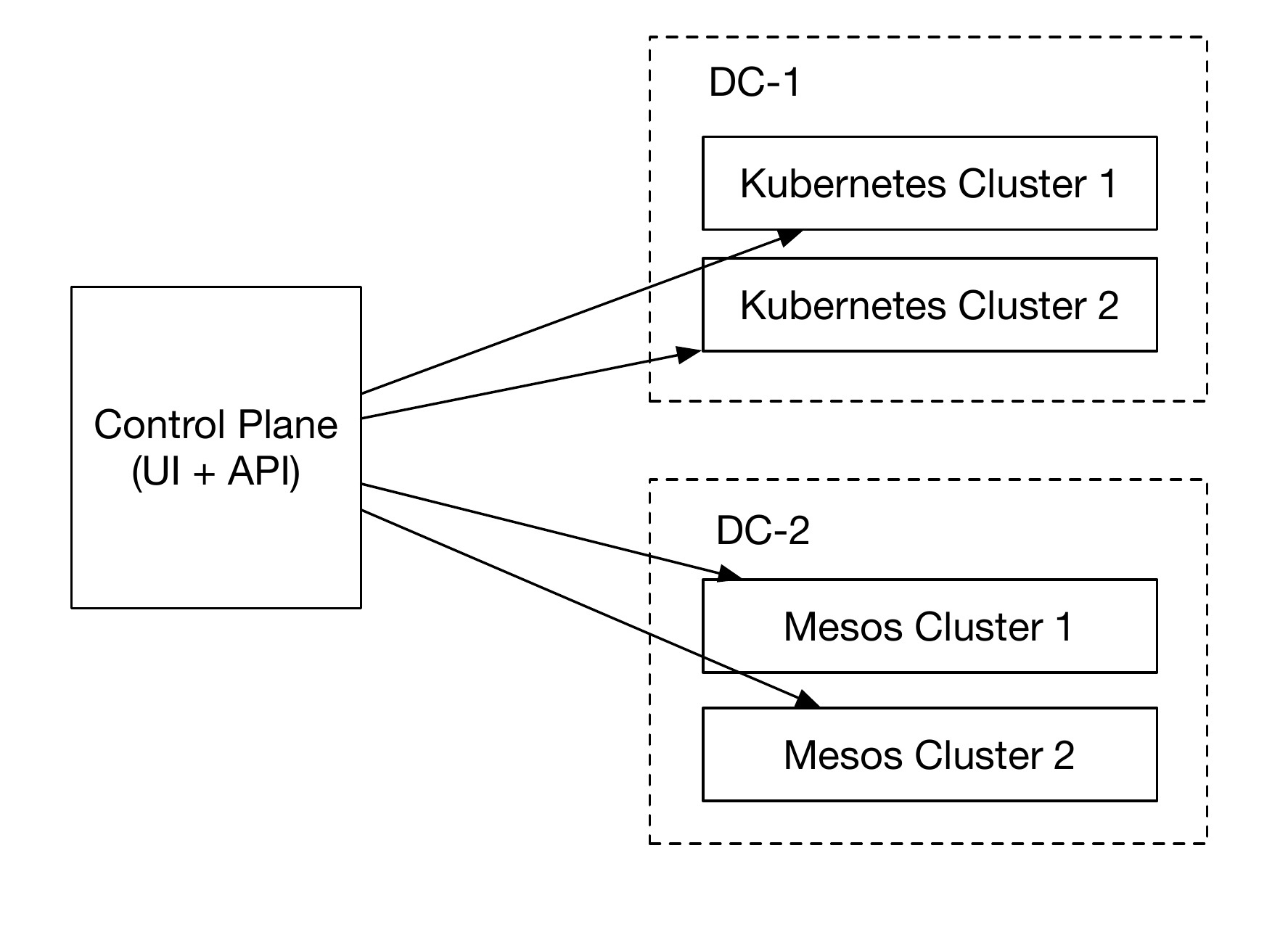}
    \vspace*{-7mm}
    \caption{Components in the UI and API layers are decoupled from container orchestration platform to form the control plane of the system, which can be deployed into a separate service platform. This makes it easy to support multiple orchestration platforms in different data centers.}
    \vspace*{-5mm}
    \label{fig:dc}
\end{figure}

One of the most important features of Alchemist is the ability to launch and manage the lifecycle of a distributed batch compute job. Alchemist is a fully containerized system. We chose Kubernetes as the container orchestration platform in the current implementation. Other than leveraging the Custom Resource Definitions (CRD) feature in Kubernetes and implementing a distributed training job resource, we took a different approach to decouple distributed job orchestration and the underlying container orchestration platform. This makes the container orchestration platform stateless and makes it possible to leverage other platforms such as Mesos in the future. Support for the cluster federation across different data centers in multiple regions is also simplified as illustrated in Figure \ref{fig:dc}.

\begin{figure}[h!]
    \includegraphics[width=\columnwidth]{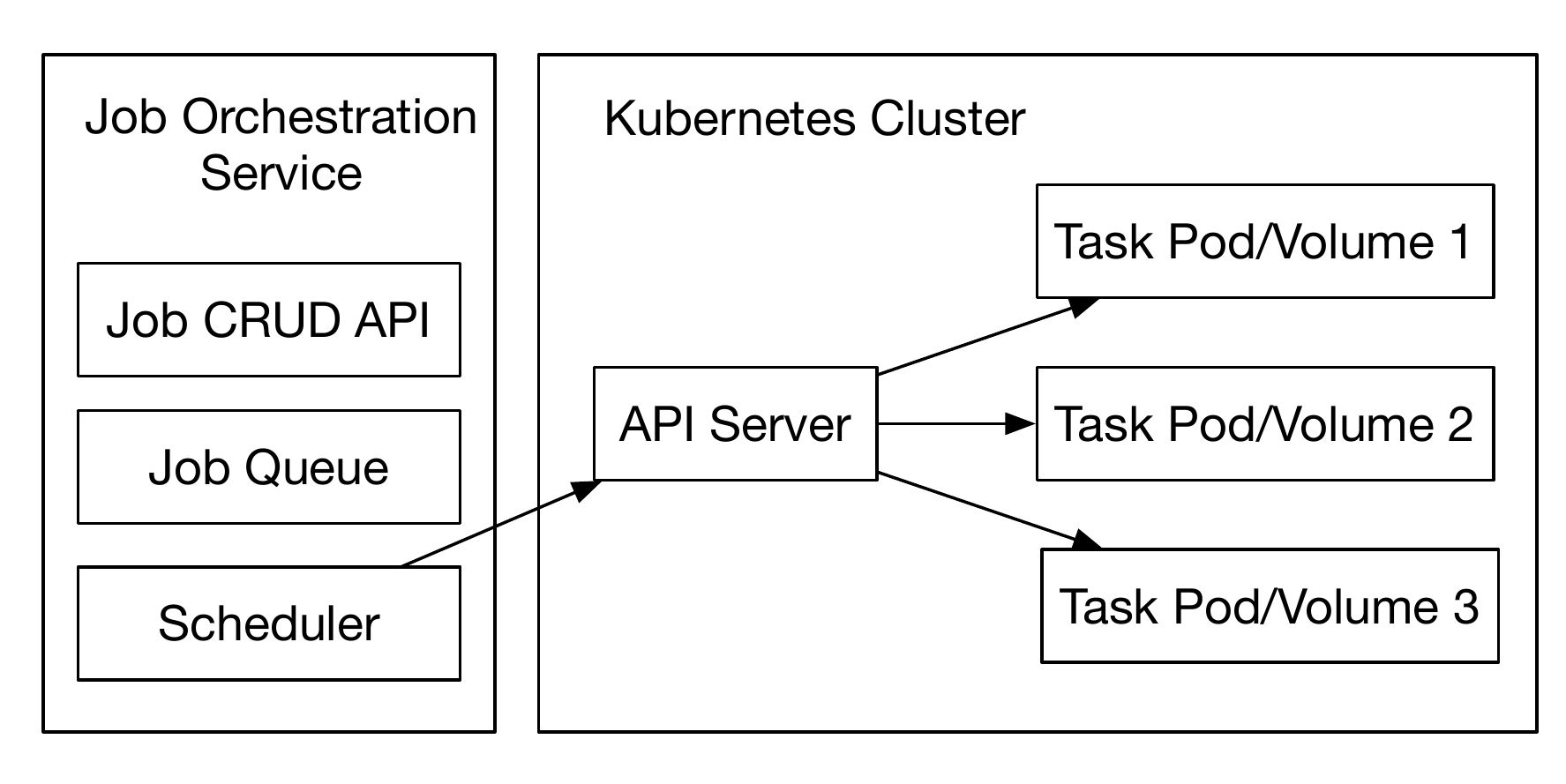}
    \vspace*{-7mm}
    \caption{Using Kubernetes as the container orchestration platform: the job scheduler launches multiple Pods for a distributed training job. It manages the life cycle of all the associated resources, e.g. storage volumes. Upon completion, the scheduler deletes all Kubernetes resources to keep the cluster stateless.}
    \vspace*{-5mm}
    \label{fig:scheduler}
\end{figure}

\textbf{Gang scheduling.} We implemented a scheduler mechanism that allocates and manages the set of tasks required for a training job. A submitted job is placed in a job queue for the scheduler to process. When the required CPU, GPU, memory, and storage resources become available, the scheduler launches the necessary task containers. It then monitors their status, and in case one fails, it will either terminate all of the job's tasks or re-launch the failed task. Upon completion of a job, the scheduler claims the free resources and resets the state in the container platform. Figure \ref{fig:scheduler} shows an example of how the job scheduler schedules a distributed training job in Kubernetes.

\begin{figure}[h!]
    \includegraphics[width=\columnwidth]{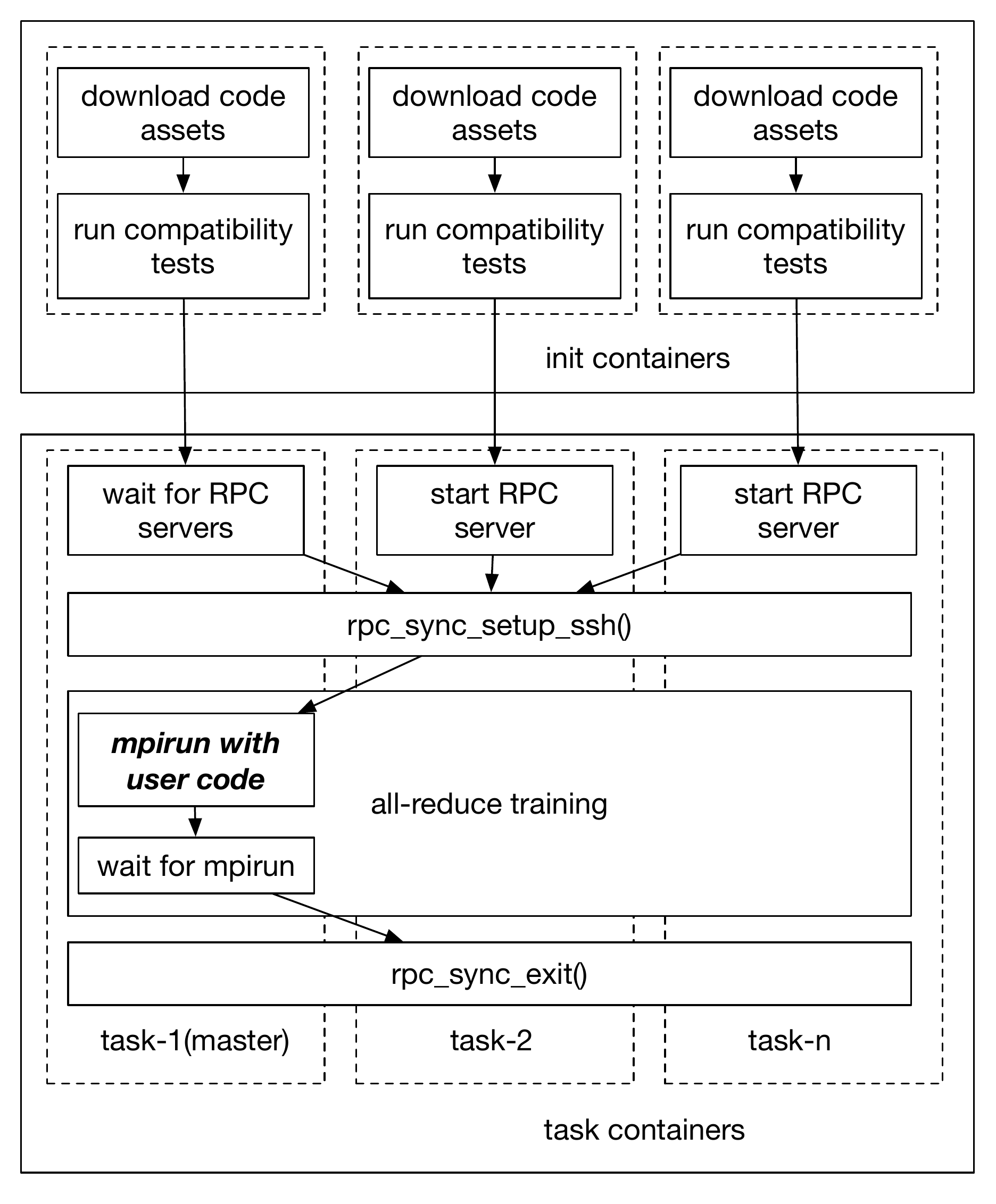}
    \vspace*{-5mm}
    \caption{Task preparation is done in separate initialization containers so that users can specify their own task images with a custom software environment. In MPI all-reduce training jobs, additional task setup is provided by the system so that users can focus on the training code without having to explicitly setup the MPI execution environment.}
    \vspace*{-5mm}
    \label{fig:harness}
\end{figure}

\textbf{Job harness.} Our job initialization step, which we call job harness, consists of two parts: a \emph{system initialization container} that runs before the user's \emph{task container}, and, optionally, code that runs inside the user's task container. The scheduler launches a system initialization container before delegating the execution to the \emph{user task container}. The initialization container downloads the user's code and runs a series of compatibility and performance tests. These tests detect common issues with the infrastructure, like networking failures and GPU driver mismatches. Making sure jobs fail fast is important for maintaining the system's performance and ease of maintenance.
Depending on the job's characteristics, we might need to execute code in the user's task container. For example, a distributed training job that uses MPI requires proper configuration of SSH credentials and cannot start until all nodes are accessible via SSH\@. We use a small remote procedure call (RPC) server to set the credentials and to synchronize all the nodes. We also use this server to gracefully stop workers upon completion. Figure \ref{fig:harness} illustrates the execution workflow.

\subsection{Storage system choices}
There are mainly four types of files stored in the system: container images, training code, datasets and model artifacts. Container images are treated separately and stored in an image registry. Users' code is stored in an object storage system. For the other two, we allow the use of either an object store or a shared filesystem. For small datasets, users can access an object storage systems by downloading the datasets before training. For medium sized datasets, a shared distributed filesystem with a POSIX compatible interface can be mounted inside task containers. For large datasets and large-scale hyper parameter tuning jobs, a high-performance in-memory storage system or streaming system is often preferred. Leveraging our gang scheduling, we are able to launch task containers on memory-optimized instances alongside GPU instances to cache and stream the datasets as the GPU instances train the model.

\subsection{Cluster auto scaling}
\begin{figure}[h!]
    \includegraphics[width=\columnwidth]{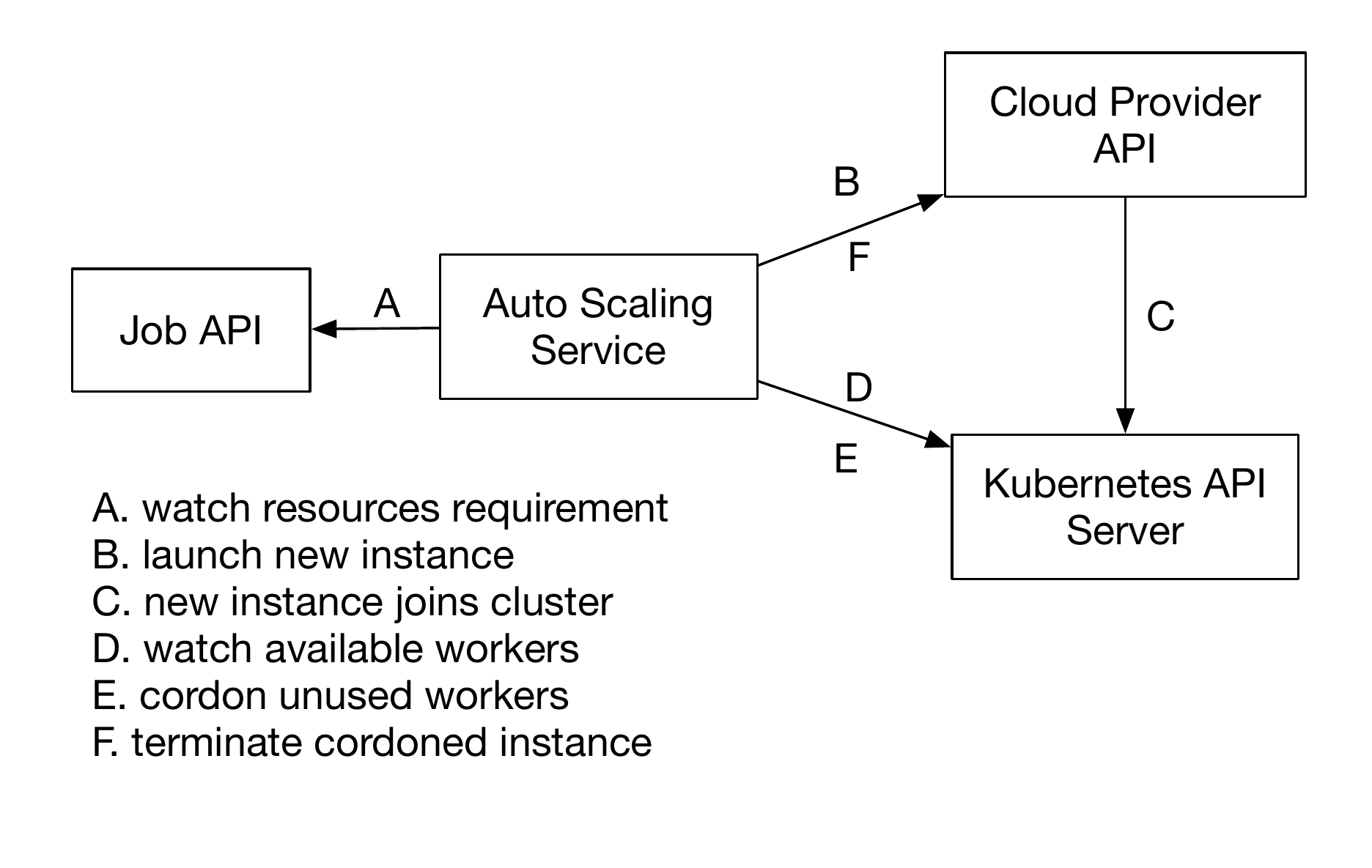}
    \vspace*{-5mm}
    \caption{The auto scaling service dynamically adjusts the size of the compute cluster by inspecting resource requests from the job queue and available workers through the container platform API (Kubernetes).}
    \vspace*{-5mm}
    \label{fig:autoscaler}
\end{figure}

Alchemist automatically scales the compute cluster based on job requests and utilization statistics. Although many private or public clouds provide out-of-the-box compute scaling based on metrics like CPU usage, a distributed training platform like Alchemist requires auto scaling on more granular metrics such as job resource requests and task status. We implemented a cluster auto scaling service optimized for distributed training as shown in Figure~\ref{fig:autoscaler}. This auto scaling service inspects the job queue for resource requirements. It uses knowledge of the available resources and launches more instances through the cloud provider API when necessary. If worker resources have been idle over a period of time, the service then terminates their host instances.

%% file: text/distributed_training.tex
\section{Distributed training in Alchemist}\label{sec:distributed_training}
\subsection{Submitting a distributed training job}
Alchemist requires users to provide a YAML configuration file to specify compute requirements. Listing \ref{lst:job} shows an example configuration, \emph{config.yaml}, for a distributed training job using Horovod with 3 GPU instances:

\lstset{language=Json, basicstyle=\ttfamily\footnotesize}
\begin{lstlisting}[float=*t, caption={Example distributed job configuration YAML.}, label=lst:job]
labels:
  name: object-detection
  optimizer: adam
spec:
  datasets_url: file-system.server:/shared/datasets/object-detection
  workspace_url: file-system.server:/object-detection/run1
  task_specs:
  - name: task-1
    image_id: tensorflow/tensorflow-gpu
    node_type: gpu-instance-type
    command: mpi-harness -np 12 -H task-1:4,task-2:4,task-3:4 python train.py
  - name: task-2
    image_id: tensorflow/tensorflow-gpu
    node_type: gpu-instance-type
    command: mpi-harness --master=false
  - name: task-3
    image_id: tensorflow/tensorflow-gpu
    node_type: gpu-instance-type
    command: mpi-harness --master=false
\end{lstlisting}

\begin{enumerate}
  \item Each job can optionally specify a read-only dataset location and a read-writable shared workspace to save model artifacts. Datasets are available inside the container filesystem when user code execution starts. The shared workspace can be used to save artifacts or synchronize data among tasks.
  \item Before the tasks are started the user's code will be downloaded into the container's filesystems. Each instance will see the code in the same location.
  \item Each task is given a unique name within a job. Tasks can communicate with each other using this name \emph{as hostname}. This avoids the need to code a mechanism for instance discovery, or worse, have hard-coded IP addresses. For example, constructing a \emph{ClusterSpec} for a distributed Tensorflow job becomes trivial.
  \item Each task can use different container images and request different types of compute instances. For example, when using parameter servers, these often require large memory and high network bandwidth whereas workers require GPUs.
  \item In a Horovod (MPI) training job, our \emph{mpi-harness} prepares SSH credentials, launches a Horovod job, and terminates workers upon completion. 
\end{enumerate}

Alchemist provides a CLI - \emph{acmctl} for users to interact with its API. To submit the above job, the incantation
\begin{verbatim}
  $ acmctl submit --config=config.yaml
    --tar=/local/code/path
\end{verbatim}
would be used.

The CLI will upload the user's code to the service and create a job in the queue, using the RESTful API. When the compute resources are available, the scheduler will launch the tasks for the job. In the meantime, users can track the job status, view job logs and metrics through the CLI or the web UI.

\subsection{Practical considerations in distributed training}
In this section, we further share our experiences of running distributed training jobs using Alchemist.

There are a few considerations we followed in order to decouple the DNN model development from the distributed training infrastructure.

\textbf{One GPU per MPI process.} We explored various assignments between GPUs and processes. Assume that there are $m$ machines, that we can have $p$ MPI processes per machine, and that we can assign $g$ GPUs to each process. That is, there are $p\cdot g$ GPUs per machine, and $m \cdot p \cdot g$ total GPUs.
For our models we found that $g=1$, i.e., one GPU per MPI process, is an optimal choice. This brings several benefits. Firstly it simplifies the ML model code. There is no need to handle multiple-GPUs inside the training code. All inter-GPU communication is handled by the distributed training framework. Since NCCL detects fast communication channels between GPUs, e.g. through NVLink, this does not lead to throughput degradation. Furthermore, splitting an 8 GPU python process into 8 single GPU processes makes better use of a multi-core architecture by lessening the effect of the global interpreter lock (GIL).

\begin{figure}[htb!]
    \includegraphics[width=\columnwidth]{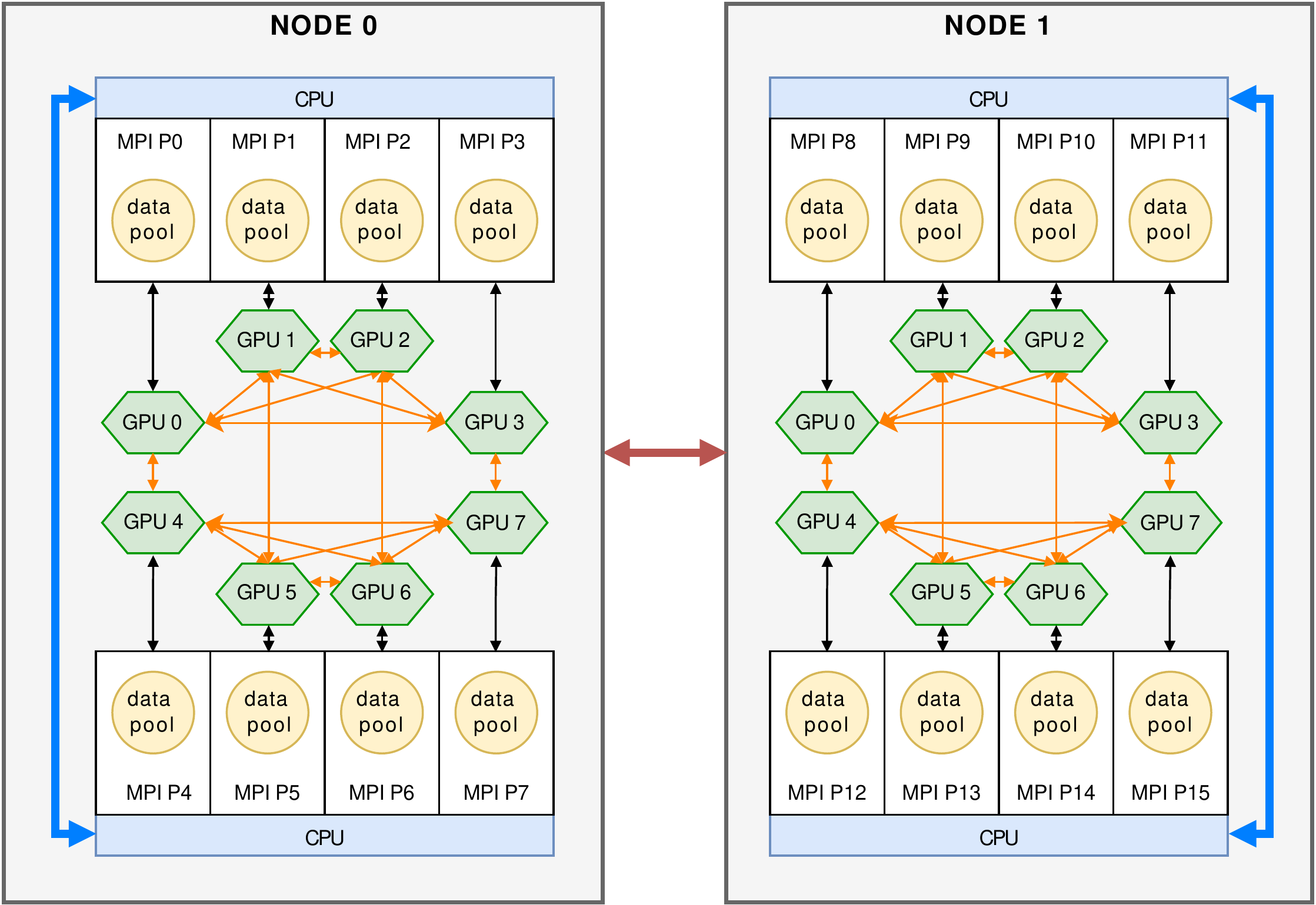}
    \vspace*{-5mm}
    \caption{GPU assignment schematic. One GPU per MPI process with one dedicated pool of pre-processed data. Several communication links (ethernet, sockets, NVLink, etc.) are shown. Optimal communication path between GPUs is handled by NCCL.}
    \vspace*{-5mm}
    \label{fig:gpu_config}
\end{figure}

For many DNN training tasks, a significant portion of time is spent on data pre-processing, e.g., image decoding, data augmentation, and forming the mini batch by shuffling and data selection. This is typically done using the CPU. We observe that splitting the process makes the data pipeline more efficient since each GPU has a process dedicated to populate its data input queue. With this configuration, moving some of the CPU pre-processing computation to the GPU does not need another scheduling since each pool of pre-processed data is assigned to exactly one GPU as demonstrated in \ref{fig:gpu_config}.

\textbf{Batch-size dependent parameters.} Many model hyper-parameters such as learning-rate scheduling, number of training steps, and evaluation frequency depend on the global batch-size. This is in turn a function of the number of GPUs used. We discovered that it's convenient to normalize all parameters to a standard batch-size, and adjust them for the particular global batch-size used in the individual the training jobs.

\textbf{Parallel evaluation.} It is common to construct training loops by alternating training and evaluation in order to monitor the model performance. In distributed settings, as training scales, more frequent evaluation is often needed. Depending on the size of the dataset and implementation details, the evaluation can occupy a large portion of the total wall time. With Alchemist, users can launch parallel evaluation tasks within a distributed training job, or launch a separate evaluation job altogether. The evaluation tasks can monitor an artifacts directory shared with the training tasks and run the evaluations in parallel. This way training tasks are not interrupted during evaluation runs.

\textbf{Distributing early.} With libraries like Horovod, implementing distributed training is simple. With Alchemist, the training infrastructure and orchestration is also simple. By adopting a containerized workflow, users can debug locally and run experiments in Alchemist remotely with the same software environment. Together, these tools enable a seamless and scalable workflow for distributed training from the beginning. In our experience adopting distributed training early reduces future integration risks and allows code to scale easily.

\subsection{Throughput scaling results}

\textbf{Experiments setup.} Unless listed otherwise, results we show below were generated on GPU servers. Each server has 64 virtual CPU cores, 488 GB memory and 8 NVIDIA Tesla V100 GPUs with peer-to-peer connection through NVLink. These servers are connected through 25GbE network. We use the Tensorflow benchmark code \footnote{\url{https://github.com/tensorflow/benchmarks}} for all experiments (git commit: \emph{9815f5f}). Major software components are NVIDIA Driver 396.26, CUDA 9.2, cuDNN 7.1, NCCL 2.2, OpenMPI 3.1.1, Tensorflow 1.9, Horovod 0.13.

\begin{figure}[!hbt]
  \includegraphics[width=70mm]{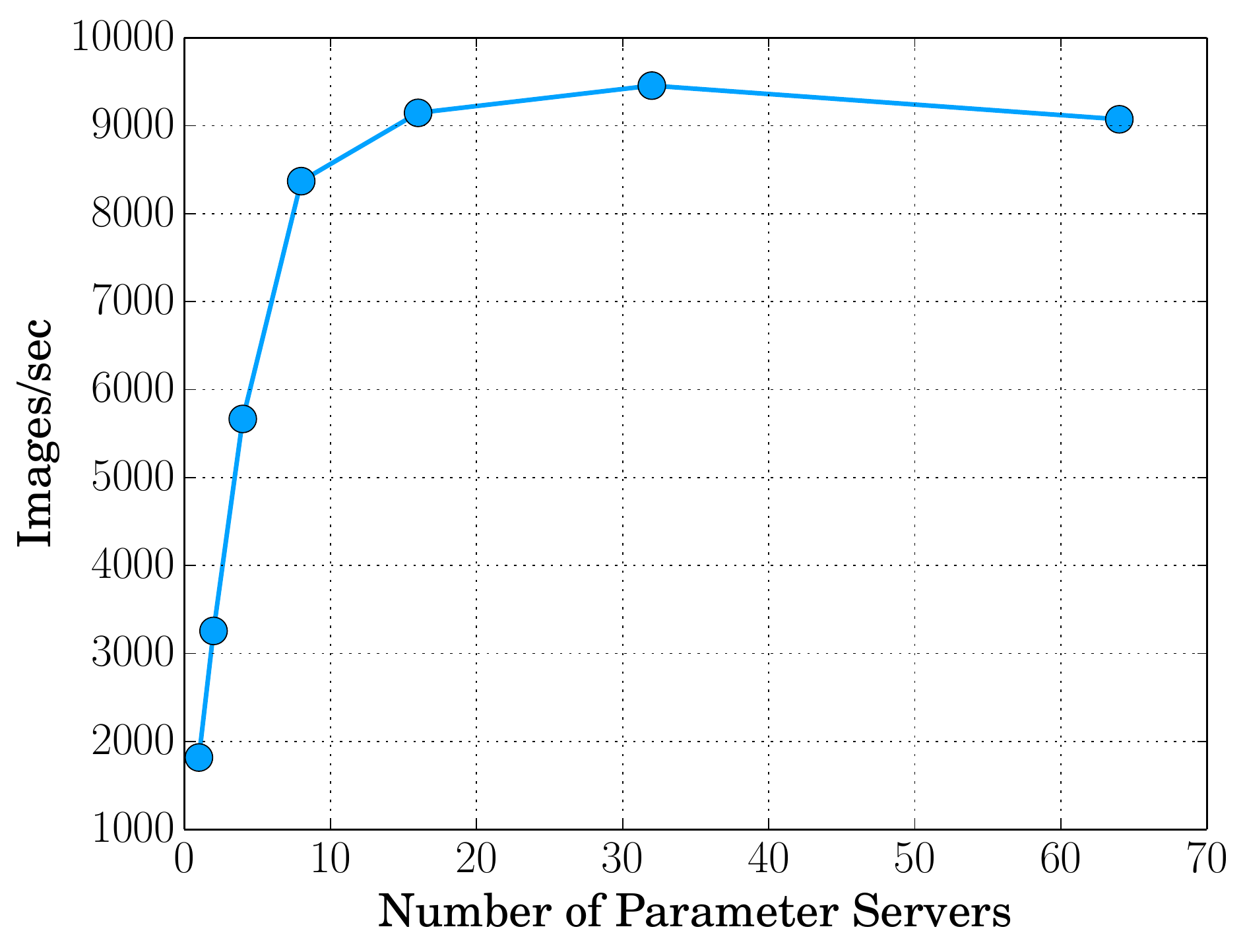}
  \vspace*{-5mm}
  \caption{Throughput vs number of parameter servers for ResNet-101 with synthesized ImageNet data. Batch size per GPU is 64. Variable update mode is \emph{parameter server}. All parameter servers are running on different hosts. Workers are fixed as 8 servers with 64 GPUs.}
  \vspace*{-2mm}
  \label{fig:ps_selection}
\end{figure}

\textbf{Results.} We first ran a naive parameter server experiment to demonstrate the capability of the system. A ResNet-101 benchmark experiment can be launched easily using the Tensorflow benchmark code and an additional Alchemist job configuration YAML. Figure \ref{fig:ps_selection} shows the scaling results with different numbers of parameter servers for a fixed number of workers with 64 GPUs. In practice, it is challenging to decide the best ratio of workers to parameter servers and the optimal placement of parameters to balance the load. 

\begin{figure*}
  \includegraphics[width=\textwidth]{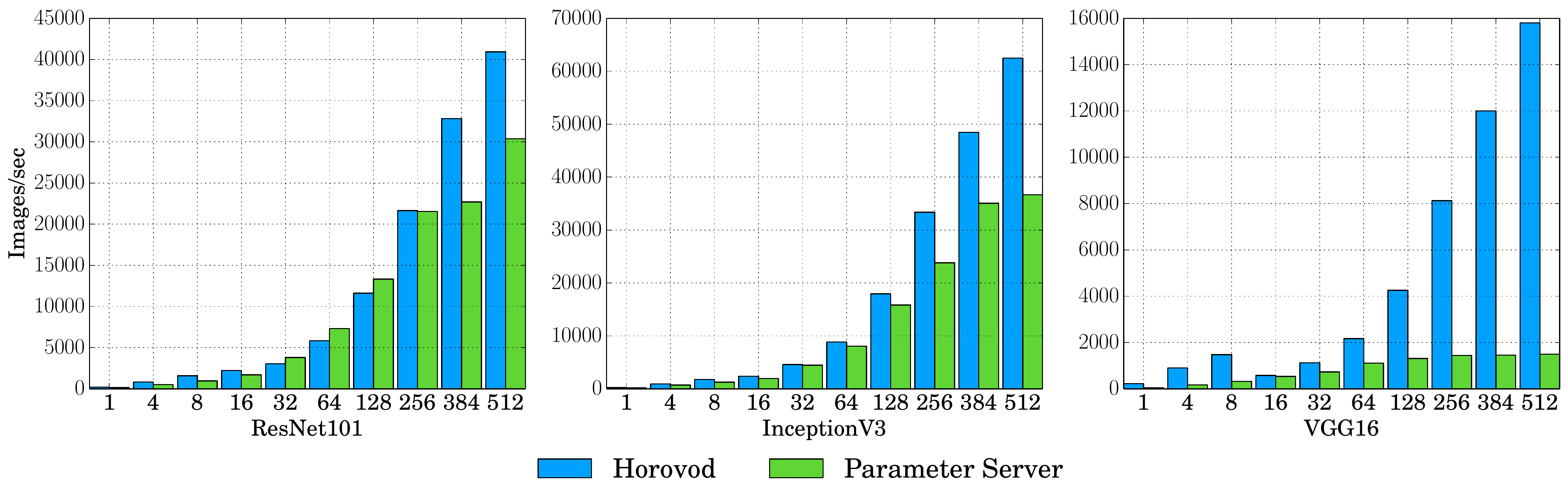}
  \vspace*{-5mm}
  \caption{Throughput vs GPUs using Horovod ring all-reduce vs parameter server for distributed training of synthesized ImageNet data for RestNet-101, Inception V3 and VGG-16. Variable update mode for parameter server is \emph{distributed replicated} and each server runs a parameter server and a worker.}
  \vspace*{-5mm}
  \label{fig:throughput_hvd_ps}
\end{figure*}

In addition, we ran experiments with different models comparing parameter server and Horovod based distributed training. To simplify the setup, we used replicated variables for parameter servers which is similar to but often better than the standard use of parameter servers \footnote{\url{https://www.tensorflow.org/performance}}. Figure \ref{fig:throughput_hvd_ps} shows the scaling results of different models as the number of GPU workers increases. In most cases, Horovod scales better than parameter server and it scales consistently well across different number of GPUs and different models. Note that parameter server scales poorly for VGG-16 due to uneven load balancing of parameter operations (\emph{fc} layers have substantially much larger weights).


%% file: text/case_studies.tex
\section{Case Study - Autonomous Systems}\label{sec:case_studies}
We now present how Alchemist has been adopted by internal teams doing research and development in autonomous systems. 

The development process of a neural network in autonomous systems consists of continuous data preparation, model training, and deployment. Especially in perception systems, due to a huge diversity of training examples and complexity of the models, DNN training requires an efficient machinery to enable rapid development cycle. This is analogous to the coding/debugging, compile, and release processes in a regular software development project. An efficient software development process requires all these components to be fast, reproducible, and traceable. Alchemist, as a powerful compute backbone for training distributed DNNs, has been one of the key components in this process.

\subsection{2D object detection in images}
\begin{figure}
    \begin{center}
    \includegraphics[width=80mm]{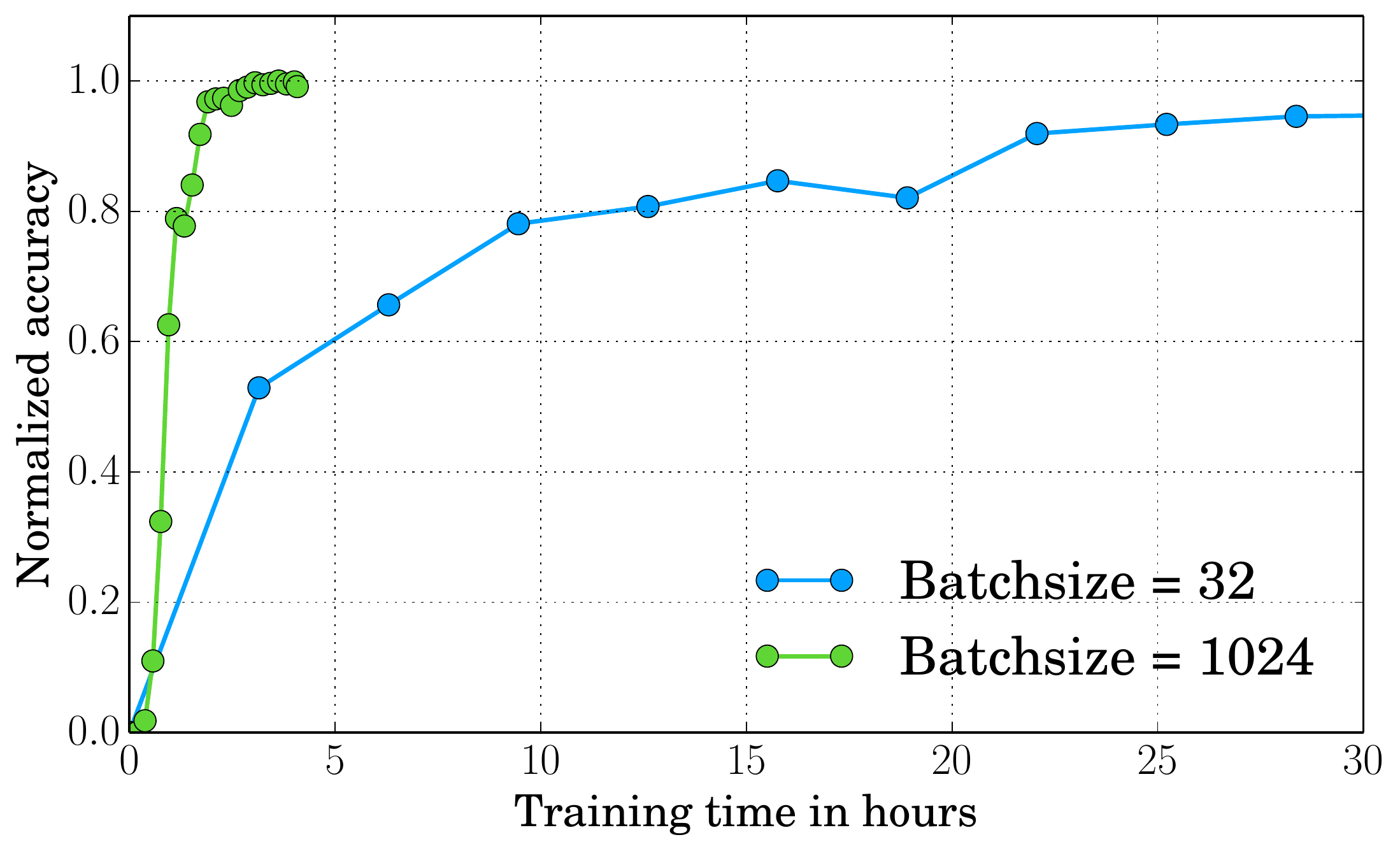}
    \end{center}
    \vspace*{-5mm}
    \caption{Convergence rate of synchronized single instance (4 GPUs, batch-size = 32) vs distributed training (64 GPUs, batch-size = 1024) for 2D object detection DNN training. Single instance training is implemented in Tensorflow across 4 GPUs. Distributed training is implemented with Horovod ring all-reduce and NCCL enabled.}
    \vspace*{-3mm}
    \label{fig:convergence}
\end{figure}

Detection and classification of 2D objects in images \cite{rpn:paper} is an essential perception task. In this case a training sample consists of an image with a set of 2D bounding-boxes that indicate the location and class of the objects. Due to the diversity of scenes and number of object types, training robust models requires very large amounts of annotated images. Using Alchemist we have been able to reduce training times from days to a few hours. This significantly accelerated the development cycle of the algorithms and allowed for the use of large-scale datasets more effectively. Figure~\ref{fig:convergence} shows an example comparing single instance training (batch-size = 32) and distributed training (batch-size = 1024). Note that as the number of GPUs is increased the global batch-size increases accordingly. This can degrade the convergence rate of batch SGD\@. To recover the original convergence, we used a scaled learning-rate (proportional to global batch-size) with an initial warm-up as suggested by \cite{goyal2017accurate}.

\begin{figure}[hbt]
    \begin{center}
    \includegraphics[width=70mm]{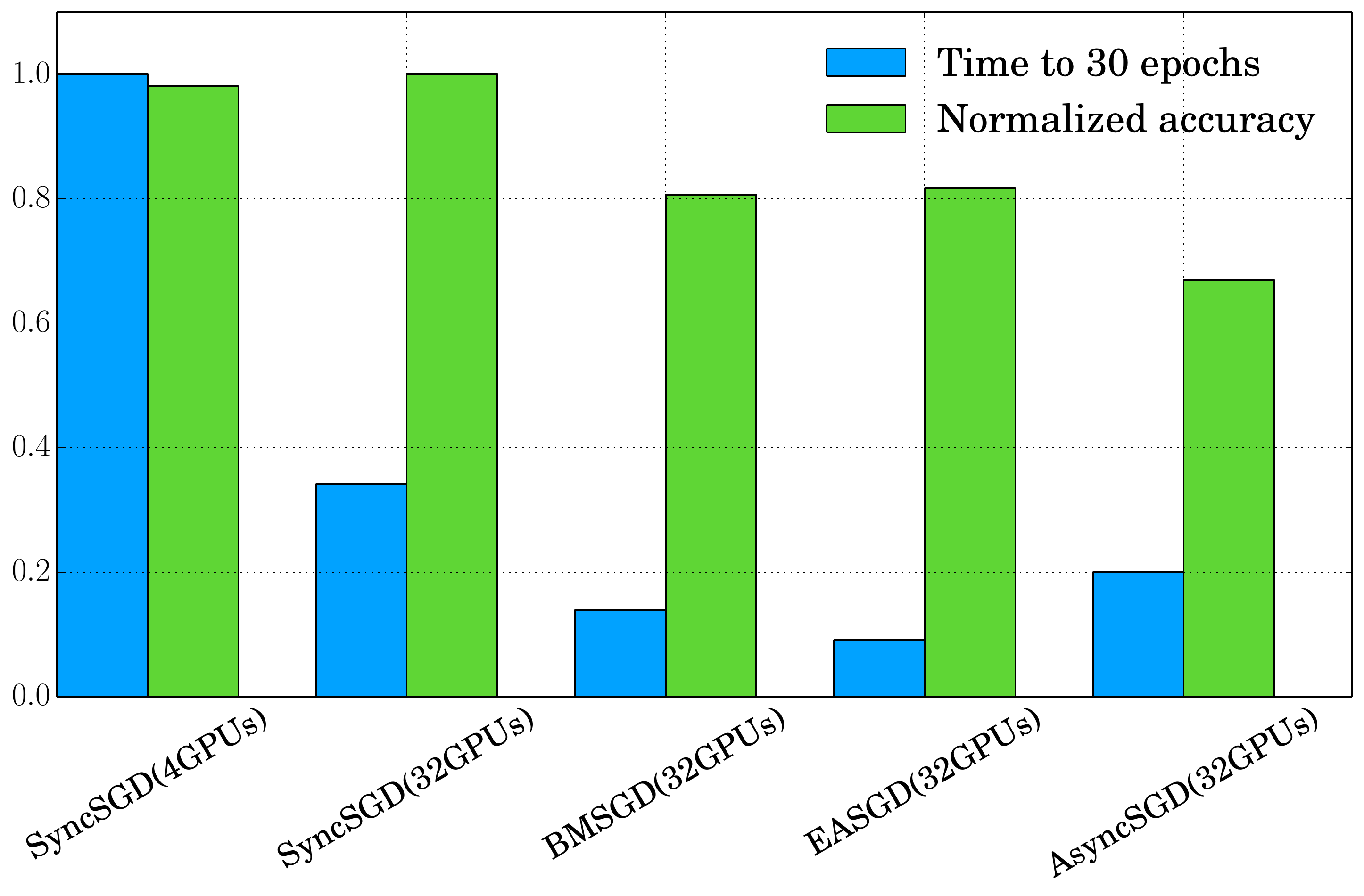}
    \end{center}
    \vspace*{-5mm}
    \caption{Comparison of throughput and convergence of 2D image detection DNN training using various distributed training algorithms: Synchronous SGD with parameter server, Block Momentum SGD \cite{bmsgd:paper}, Elastic Averaging SGD \cite{easgd:paper}, Asynchronous SGD.}
    \vspace*{-3mm}
    \label{fig:rpn_distalg_comparison}
\end{figure}

Fully synchronous SGD is not the only distributed training algorithm. To explore our options we ran several experiments with other asynchronous variants and validated that (for the case of our 2D object detection models) our choice of fully synchronous distributed SGD achieved low wall-time and highest accuracy. As shown in Figure \ref{fig:rpn_distalg_comparison} we observe that the asynchronous algorithms (asynchronous SGD, Block Momentum SGD, Elastic Averaging SGD) have better throughput but converge more slowly than synchronous algorithms.

\subsection{3D object detection in point cloud}
Besides cameras, another commonly used sensor in autonomous perception systems is LiDAR. It provides rich depth information used by applications like 3D mapping and 3D object detection. Recent work like \cite{voxelnet:paper} proposes an end-to-end trainable network for 3D object detection. Similar to 2D object detection, with minimal modifications to single-machine training code we were able to use Alchemist and distribute VoxelNets training. Using 64 GPUs (on 8 machines) we were able to reduce the training time by a factor of 14 w.r.t our baseline, which uses 4 GPUs on a single machine. We must reiterate that the model's accuracy is also preserved when large batches are used, as long as the learning rate schedule is corrected appropriately.

\subsection{Semantic segmentation}
Another common task for perception systems is semantic segmentation. In this task each pixel in an image receives the label of its semantic class. With Alchemist we were able to train distributed image segmentation models easily. Porting our training job from a single machine with 4 GPUs to a distributed training setup with 64 GPUs, with no further runtime optimization, we were able to reduce the training time by a factor of 11, again with no accuracy loss.

%% file: text/future_work.tex
\section{Future Work and Conclusion}\label{sec:future_work}

\subsection{Convergence of training with very large batches}
\begin{figure}[!hbt]
\begin{center}
\includegraphics[width=70mm]{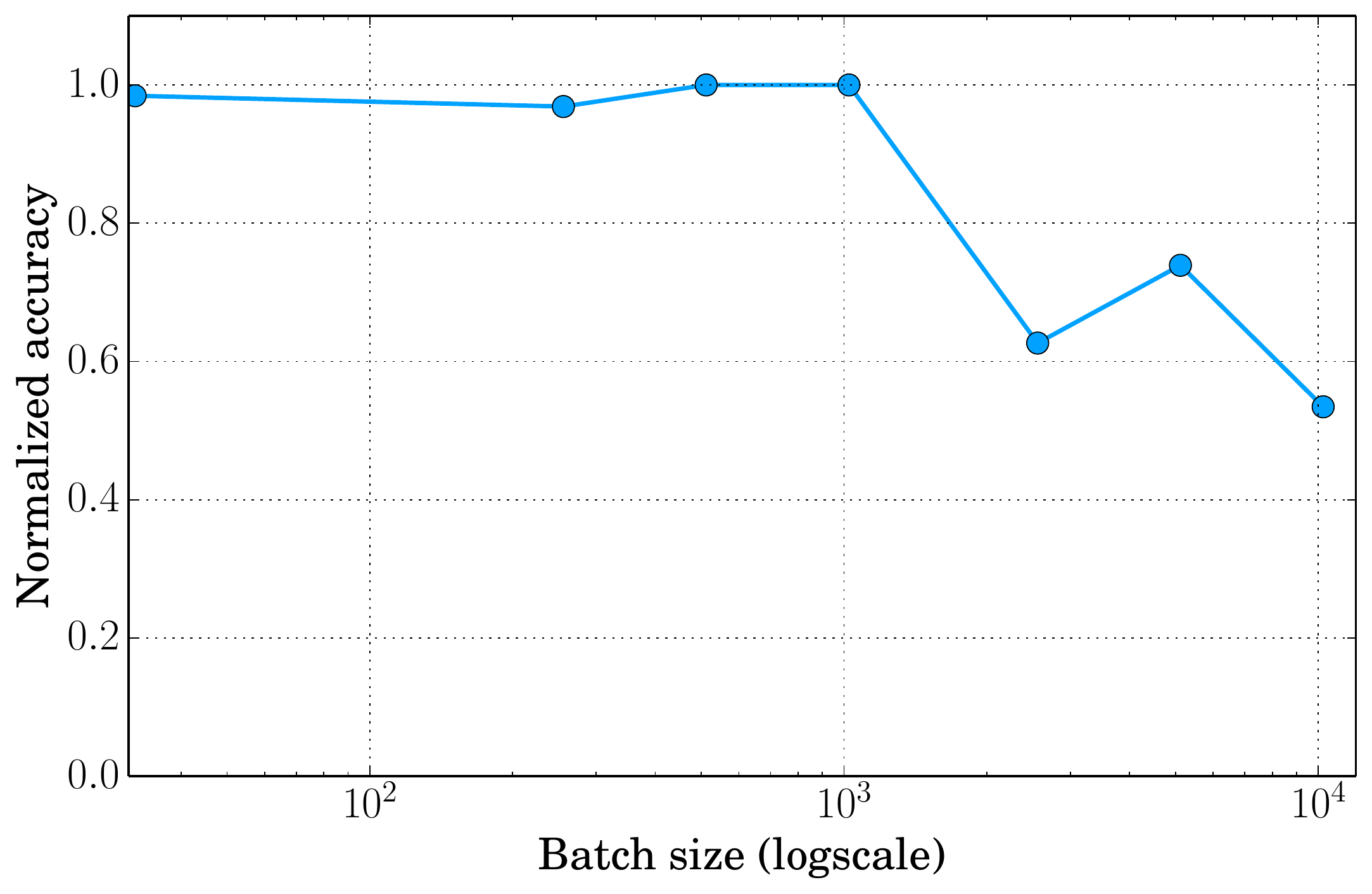}
\end{center}
\vspace*{-5mm}
\caption{Batch size limit of synchronous DNN training. After batch size is increased to a certain size, the accuracy of the model starts degrading for the same number of training epochs.}
\vspace*{-5mm}
\label{fig:largebatch}
\end{figure}

To leverage distributed training with synchronous SGD one must compensate for the large global batch size. This is done in practice by adjusting the learning rates appropriately \cite{goyal2017accurate}. We ran experiments to determine the limits of this large batch training strategy and observed that  
(for our 2D image detection model) at about batch sizes of $10^3$ we were unable to maintain the accuracy by simply adjusting the learning rates. See Figure \ref{fig:largebatch} for a plot of our results.
We intend to devise better algorithms to further scale distributed training, to this end we follow work like \cite{lars:paper,tencent4min} closely.

\subsection{Hyper parameter tuning}
In a typical DNN workflow, hyper parameter tuning is a vital step. Naive searches over the hyper parameter space (grid or random) are commonly used but often wasteful. On the other hand Bayesian Optimization \cite{bayesian_opt:paper} techniques can be used to reduce the total computation required for the same accuracy \cite{vizier:paper}. We intend to support both the naive search strategies together with a Bayesian optimization strategy. 

\subsection{Workflow management}
Utilizing deep neural networks involves a pipeline consists of data preparation, model training, and deployment. It is desired to automate the workflow on this pipeline in order to increase efficiency, clarity, and traceability. Solutions such as Airflow \footnote{\url{https://airflow.apache.org}} have already been introduced to automate workflow pipelines represented as a directed acyclic graph (DAG) of tasks. We plan to integrate Alchemist into such workflow management systems.

\subsection{Conclusion}
We presented the motivation, design and implementation of Alchemist: a service to enable easy, fast, and scalable distributed training for multiple teams within Apple. Its value has been proven in developing autonomous systems. We expect the adoption to grow further as we explore more in the future work mentioned above.